\documentclass{article}
\usepackage{stywhispers,amsmath,amssymb,epsfig,pgfplots,cite}
\usepackage[skip=8pt]{caption}
\usetikzlibrary{backgrounds}

\title{Multilayer Structured NMF for Spectral Unmixing of Hyperspectral Images}
\name{Roozbeh Rajabi, Hassan Ghassemian} 
\address{ECE Department, Tarbiat Modares University, Tehran, Iran}
\begin{document}
%
\maketitle
\begin{abstract}
One of the challenges in hyperspectral data analysis is the presence of mixed pixels. Mixed pixels are the result of low spatial resolution of hyperspectral sensors. Spectral unmixing methods decompose a mixed pixel into a set of endmembers and abundance fractions. Due to nonnegativity constraint on abundance fraction values, NMF based methods are well suited to this problem. In this paper multilayer NMF has been used to improve the results of NMF methods for spectral unmixing of hyperspectral data under the linear mixing framework. Sparseness constraint on both spectral signatures and abundance fractions matrices are used in this paper. Evaluation of the proposed algorithm is done using synthetic and real datasets in terms of spectral angle and abundance angle distances. Results show that the proposed algorithm outperforms other previously proposed methods.
\end{abstract}
\section{Introduction}
\label{sec:intro}

\begin{figure}[!t]
	\centering
	\centerline{\epsfig{figure=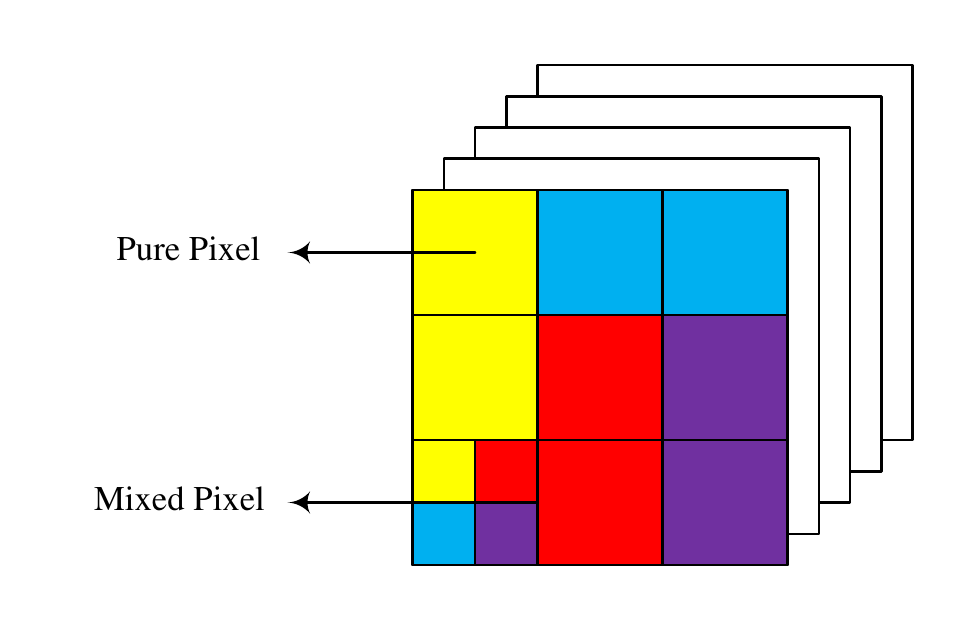,width=6.5cm}}
	\caption{Toy example of mixed pixels concept in hyperspectral images \cite{MVIP11_GNMF}.}
	\label{fig:mixed}
	\vspace{-1.5em}
\end{figure}

Hyperspectral sensors are an effective tool for remote sensing applications. These sensors offer high spectral resolution, so they can gather a lot of information about the spectral content of materials that are present in the scene. But the main drawback of these sensors is the low spatial resolution that cause mixed pixels to appear in hyperspectral images \cite{SPM14_Perspective}. The concept of mixed pixels in hyperspectral images is illustrated in Fig. \ref{fig:mixed} using a toy example.

There are three different categories of spectral unmixing methods: geometrical, statistical and sparse regression methods \cite{MLNMF:JSTARS12_Overview}. Vertex component analysis (VCA) \cite{TGRS05_VCA} is one of the classic methods in the geometrical category that is used in this paper as a baseline of the proposed method.

Another class of algorithms that are used for spectral unmixing purposes are methods based on nonnegative matrix factorization (NMF). NMF method gets a lot of attention in this area of research because of nonnegativy constraint on abundance fraction values. Different constraints can be used in NMF methods, for example in \cite{TGRS07_MVC-NMF} minimum volume constraint is used or in \cite{IGARSS13_sparseGNMF} graph regularized constraint has been used. In this paper multilayer NMF \cite{EL06_MLNMF} is used to unmix hyperspectral data. In this approach the observation matrix will be decomposed in different layers. So the final estimated spectral signatures can be modeled as a product of sparse matrices obtained in the layers. The reason why decomposing a general matrix into sparse matrices will improve the results of NMF method is still an open problem \cite{ANCA07_MLNMF}. Experiments on synthetic and real datasets show that the proposed method can achieve better results in comparison with other methods.

The rest of the paper are as follows. Section \ref{sec:LMM} introduces mathematical description of linear mixing model for solving spectral unmixing problem. Section \ref{sec:NMF} briefly reviews NMF method and commonly used sparseness constraint on abundance fractions matrix. In section \ref{sec:MLNMF} multilayer structured NMF has been described to use in hyperspectral unmixing. Experiments
on synthetic and real datasets has been done to evaluate the proposed method in section \ref{sec:evaluation}. Finally section \ref{sec:conclusion} concludes the paper.

\section{Linear Mixing Model (LMM)}
\label{sec:LMM}
Spectral unmixing problem can be solved under linear or nonlinear frameworks \cite{SPM14_Nonlinear}. In this paper linear mixing model (LMM) has been used. The mathematical representation of this model is expressed in (\ref{lmm}).
\begin{equation}
\label{lmm}
\mathbf{X = AS+N},
\end{equation}
where $\mathbf{X}\in{\mathbb{R}^{L\times{N}}}$  refers to observation matrix, $L$ denotes number of spectral bands and N denotes total number of pixels. $\mathbf{A}\in{\mathbb{R}^{L\times{P}}}$ and $\mathbf{S}\in{\mathbb{R}^{P\times{N}}}$ refer to signature and abundance fractions matrices respectively. $P$ denotes number of pure signatures (called endmembers) in the scene and  $\mathbf{N}\in{\mathbb{R}^{L\times{N}}}$  refers to measurement noise.
Two constraints on abundance fraction values should be considered: Abundance Nonnegativity Constraint (ANC) and Abundance Sum to one Constraint (ASC). These constraints are expressed in (\ref{anc}) and (\ref{asc}).
  \begin{equation}
  \label{anc}
  \forall{i,j}:\mathbf{S}_{ij}\geq{0},
  \end{equation}
  \begin{equation}
  \label{asc}
  \sum_{i=1}^P{\mathbf{S}_{ij}=1}.
  \end{equation}

\section{Nonnegative Matrix Factorization (NMF)}
\label{sec:NMF}

Nonnegative matrix factorization can decompose a matrix into a product of two matrices efficiently. NMF can be expressed mathematicaly as:

\begin{equation}
	\label{NMF}
	\mathbf{X \approx AS}.
\end{equation}

To calculate this approximation, the following cost function can be defined using Euclidean distance \cite{ANIPS00_AlgorithmsNMF}:

\begin{equation}
\label{onmf}
\mathcal{O}_{NMF}=\lVert{\mathbf{X-AS}}\rVert_{F}^{2}
\end{equation}

As explained in section \ref{sec:intro}, due to ANC on abundance fractions matrix, NMF methods has been widely used for spectral unmixing of hyperspectral data. Without employing other constraints to NMF method, the results are not satisfactory. One of the most common constraints is sparseness constraint on abundance fraction matrix \cite{MLNMF:TGRS11_L05NMF}. Considering this constraint the cost function of NMF method changes to the following equation:

\begin{equation}
\label{osparsenmf}
\mathcal{O}_{Sparse NMF}=\lVert{\mathbf{X-AS}}\rVert_{F}^{2}+\lambda\lVert{\mathbf{S}}\rVert_{1/2},
\end{equation}

\section{Multilayer NMF}
\label{sec:MLNMF}
Cichocki and Zdunek \cite{EL06_MLNMF} proposed multilayer structure for NMF to reduce the risk of getting stuck in local minima and improve the performance of NMF methods. In this paper we have used this idea to solve spectral unmixing of hyperspectral data.

Using multilayer structure, the first layer of algorithm is a basic decomposition of observation matrix into $\mathbf{A}_1$ and $\mathbf{S}_1$ matrices. In the second layer $\mathbf{S}_1$ will be decomposed into $\mathbf{A}_2$ and $\mathbf{S}_2$ and this process will be repeated to reach the maximum number of layers. Then the estimated spectral signature matrix can be calculated using (\ref{multilayer}).

\begin{equation}
\label{multilayer}
\mathbf{A}=\mathbf{A}_1\mathbf{A}_2\hdots\mathbf{A}_{L_{max}},
\end{equation}
where $L_{max}$ is the maximum number of layers.

Considering the sparseness constraint on spectral signatures and abundance fraction matrices, the proposed cost function for each layer is expressed in (\ref{omlnmf}).
\begin{equation}
\label{omlnmf}
\mathcal{O}_{MLNMF}=\lVert{\mathbf{X-AS}}\rVert_{F}^{2}+\alpha\lVert{\mathbf{A}}\rVert_{1/2}+\lambda\lVert{\mathbf{S}}\rVert_{1/2},
\end{equation}
where $\alpha$ and $\lambda$ are regularization parameters and control the impact of sparseness constraints on spectral signatures and abundance fractions respectively.

Optimizing the cost function in \ref{omlnmf} using multiplicative rules results in estimated $\mathbf{A}$ and $\mathbf{S}$. Note that the cost function in (\ref{omlnmf}) should be minimized in each level of the algorithm. In this paper to consider the ASC constraint on abundance fraction values, FCLS method \cite{TGRS01_FCLS} has been used.

\section{Evaluation Results}
\label{sec:evaluation}
To evaluate the performance of the proposed algorithm, spectral angle distance (SAD) and abundance angle distance (AAD) are used. These metrics are defined in (\ref{sad}) and (\ref{aad}).

\begin{equation}
\label{sad}
{SAD}_{m_{i}}=\cos^{-1}{(\frac{\mathbf{m}_{i}^{T}\hat{\mathbf{m}}_{i}}{\lVert{\mathbf{m}_{i}}\rVert\lVert{\hat{\mathbf{m}}_{i}}\rVert})}
\end{equation}

\begin{equation}
\label{aad}
AAD_{a_{i}}=\cos^{-1}{(\frac{\mathbf{a}_{i}^{T}\hat{\mathbf{a}}_{i}}{\lVert{\mathbf{a}_{i}}\rVert\lVert{\hat{\mathbf{a}}_{i}}\rVert})}
\end{equation}

SAD and AAD are defined for one endmember and one pixel respectively. To have an overall measure for all endmembers and pixels, root mean square values of these metrics have been used in this paper and defined in (\ref{rmssad}) and (\ref{rmsaad}).

\begin{equation}
\label{rmssad}
rmsSAD=(\frac{1}{P}\sum\limits_{i=1}^{P}({\text{SAD}_{\mathbf{m}_{i}}})^{2})^{1/2}
\end{equation}
\begin{equation}
\label{rmsaad}
rmsAAD=(\frac{1}{N}\sum\limits_{i=1}^{N}({\text{AAD}_{\mathbf{a}_{i}}})^{2})^{1/2}
\end{equation}

\subsection{Synthetic Dataset}
\label{ssec:synthetic}
\begin{figure}[!t]
   	\newlength\figureheight
   	\newlength\figurewidth
   	\setlength\figureheight{4cm}
   	\setlength\figurewidth{7.5cm}
   	\input{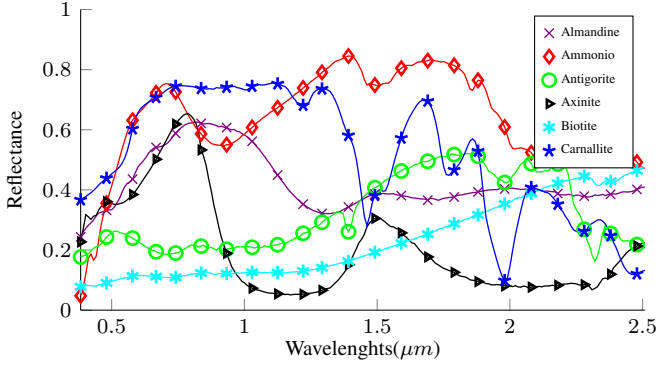}
   	\caption{Selected Materials from USGS Library.}
   	\label{fig_SelectedMats}
   \end{figure}

In this experiment, synthetic data has been generated using spectral signatures of USGS spectral library \cite{USGS07_splib06}. Selected materials from this library are shown in Fig. \ref{fig_SelectedMats}. The process described in \cite{TGRS07_MVC-NMF} has been used in this paper to generate simulated data.

\begin{figure}[!t]
   	\centering
   	\setlength\figureheight{5cm}
   	\setlength\figurewidth{5cm}
%
%

\begin{center}
\begin{tikzpicture}[show background rectangle, tight background] 

\begin{axis}[%
name=plot1,
width=\figurewidth,
height=\figureheight,
scale only axis,
xmin=15,
xmax=45,
xlabel style={yshift=0.25em, font=\small},
xticklabel style={font=\footnotesize},
xlabel={SNR (dB)},
ymin=0,
ymax=0.25,
yticklabel style={font=\footnotesize},
ylabel style={yshift=-0.3em,  font=\small},
ylabel={rmsSAD},
legend style={draw=black,fill=white,legend cell align=left, font=\footnotesize, mark options={scale=2}},
]
\addplot [color=blue,solid,line width=1pt,mark size=2pt,mark=diamond,mark options={solid}]
  table[row sep=crcr]{
15	0.10590453767967	\\
20	0.0538640312528068	\\
25	0.0333021475841421	\\
30	0.0367763487303866	\\
35	0.0359891688013195	\\
40	0.0347751581165641	\\
45	0.0357307384215763	\\
};
\addlegendentry{MLNMF};

\addplot [color=red,dotted,line width=1pt,mark size=1.5pt,mark=square,mark options={solid}]
  table[row sep=crcr]{
15	0.150475408848932	\\
20	0.0992352917924799	\\
25	0.0887563010125809	\\
30	0.0850839503897561	\\
35	0.0826517599094664	\\
40	0.0805193784120333	\\
45	0.0840594545758741	\\
};
\addlegendentry{$\text{L}_{\text{1/2}}\text{-NMF}$};

\addplot [color=green,dash pattern=on 1pt off 3pt on 3pt off 3pt,line width=1pt,mark size=2pt,mark=o,,mark size=2pt,mark options={solid}]
  table[row sep=crcr]{
15	0.167713489438206	\\
20	0.172767435277796	\\
25	0.164925879117054	\\
30	0.161643542282438	\\
35	0.173944955191514	\\
40	0.180254304398925	\\
45	0.163325338137687	\\
};
\addlegendentry{VCA};
\end{axis}
\end{tikzpicture}
\\
\begin{tikzpicture}[show background rectangle, tight background] 
\begin{axis}[%
name=plot2,
width=\figurewidth,
height=\figureheight,
scale only axis,
xmin=15,
xmax=45,
xlabel style={yshift=0.25em, font=\small},
xticklabel style={font=\footnotesize},
xlabel={SNR (dB)},
ymin=0,
ymax=0.7,
yticklabel style={font=\footnotesize},
ylabel style={yshift=-0.3em, font=\small},
ylabel={rmsAAD},
legend style={draw=black,fill=white,legend cell align=left, font=\footnotesize} 
]
\addplot [color=blue,solid,line width=1pt,mark size=2pt,mark=diamond,mark options={solid}]
  table[row sep=crcr]{
15	0.335746733189388	\\
20	0.198415707894966	\\
25	0.133089513322697	\\
30	0.124303868944505	\\
35	0.132455505955305	\\
40	0.121830738056306	\\
45	0.126954431892322	\\
};
\addlegendentry{MLNMF};

\addplot [color=red,dotted,line width=1pt,mark size=1.5pt,mark=square,mark options={solid}]
  table[row sep=crcr]{
15	0.48705322337738	\\
20	0.335355245059084	\\
25	0.305940553708276	\\
30	0.27078055089705	\\
35	0.2693405605796	\\
40	0.263741861798504	\\
45	0.25238171981101	\\
};
\addlegendentry{$\text{L}_{\text{1/2}}\text{-NMF}$};

\addplot [color=green,dash pattern=on 1pt off 3pt on 3pt off 3pt,line width=1pt,mark size=2pt,mark=o,mark options={solid}]
  table[row sep=crcr]{
15	0.521922201011502	\\
20	0.509700361135389	\\
25	0.403648244856564	\\
30	0.474222468094169	\\
35	0.440086151529465	\\
40	0.519957240601344	\\
45	0.481951899103929	\\
};
\addlegendentry{VCA};
\end{axis}
\end{tikzpicture}%
\end{center}
   	\caption{Comparison of Results in terms of  rmsSAD (Top) and rmsAAD (Bottom) for VCA,  $L_{1/2}$-NMF and MLNMF against SNR.}
   	\label{fig_rmsSADrmsAAD}
   	\vspace{-1.5em}
   \end{figure}
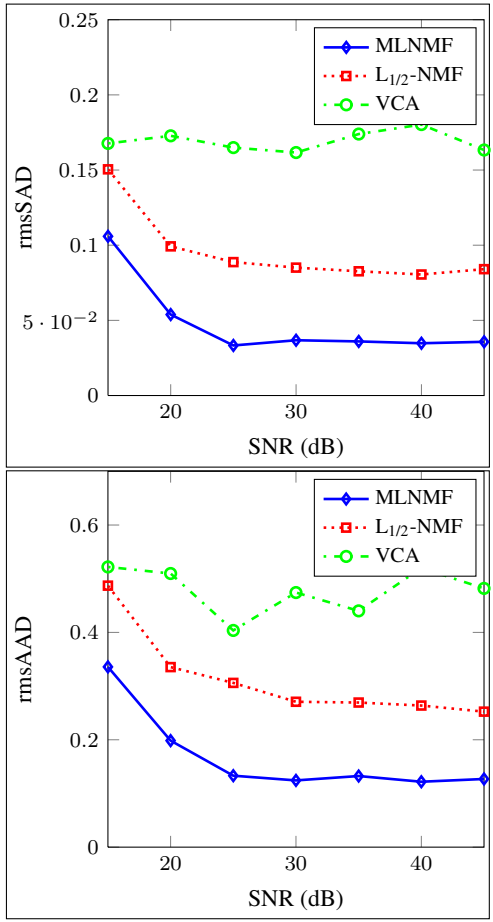

Fig. \ref{fig_rmsSADrmsAAD} shows the results of applying proposed method on synthetic dataset. This figure illustrates the results in terms of rmsSAD and rmsAAD in comparison with VCA \cite{TGRS05_VCA} and $L_{1/2}$-NMF \cite{MLNMF:TGRS11_L05NMF} for different values of SNR ranging from $15dB$ to $45dB$. As it can be seen from results the proposed method can unmix data more effectively compared to other methods.

\subsection{Real Dataset}
\label{ssec:real}
In the second experiment Cuprite Nevada dataset collected by AVIRIS sensor \cite{MLNMF:AVIRIS_Cuprite} is used as real dataset for evaluating the proposed method. This dataset contains $250\times191$ pixels and $188$ spectral bands. This area consists of different minerals. Reference spectra to compare the results are selected from USGS library \cite{USGS07_splib06}. 

\begin{figure}[!t]
   	\centering
   	\setlength\figureheight{1.25cm}
   	\setlength\figurewidth{1.75cm}
   	\input{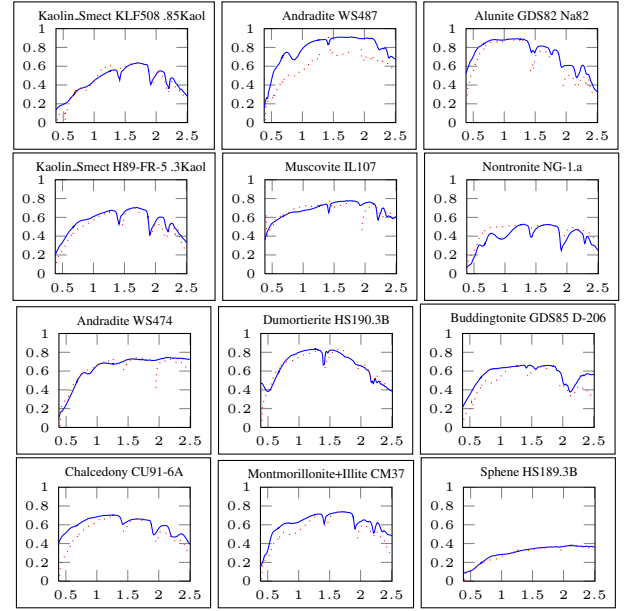}
   	\caption{Estimated signatures (dotted in red) in comparison with USGS library signatures (continous in blue) for Cuprite dataset}
   	\label{fig_cupriteSignatures}
   	\vspace{-1.5em}
   \end{figure}

\begin{figure}[!t]
   	\centering
   	\setlength\figureheight{2.6cm}
   	\setlength\figurewidth{2.25cm}
%
%
\pgfplotsset{every axis title/.append style={at={(0.5,0.9)}}}
\begin{tikzpicture}[show background rectangle, tight background, font=\tiny]
\begin{axis}[%
width=\figurewidth,
height=\figureheight,
axis on top,
scale only axis,
xmin=0.5,
xmax=191.5,
y dir=reverse,
ymin=0.5,
ymax=250.5,
hide axis,
name=plot1,
title = {Alunite GDS82 Na82 }
]
\addplot [forget plot] graphics [xmin=0.5,xmax=191.5,ymin=0.5,ymax=250.5] {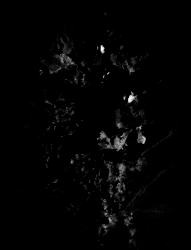};
\end{axis}
\end{tikzpicture}
\begin{tikzpicture}[show background rectangle, tight background, font=\tiny]
\begin{axis}[%
width=\figurewidth,
height=\figureheight,
axis on top,
scale only axis,
xmin=0.5,
xmax=191.5,
y dir=reverse,
ymin=0.5,
ymax=250.5,
hide axis,
name=plot2,
title={Andradite WS487 Garnet }
]
\addplot [forget plot] graphics [xmin=0.5,xmax=191.5,ymin=0.5,ymax=250.5] {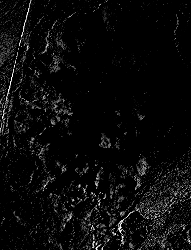};
\end{axis}
\end{tikzpicture}
\begin{tikzpicture}[show background rectangle, tight background, font=\tiny]
\begin{axis}[%
width=\figurewidth,
height=\figureheight,
axis on top,
scale only axis,
xmin=0.5,
xmax=191.5,
y dir=reverse,
ymin=0.5,
ymax=250.5,
hide axis,
name=plot3,
title={Buddingtonite GDS85 D-206 }
]
\addplot [forget plot] graphics [xmin=0.5,xmax=191.5,ymin=0.5,ymax=250.5] {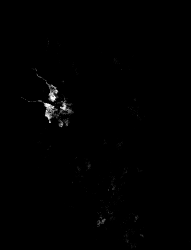};
\end{axis}
\end{tikzpicture}
\\
\begin{tikzpicture}[show background rectangle, tight background, font=\tiny]
\begin{axis}[%
width=\figurewidth,
height=\figureheight,
axis on top,
scale only axis,
xmin=0.5,
xmax=191.5,
y dir=reverse,
ymin=0.5,
ymax=250.5,
hide axis,
name=plot4,
title={Chalcedony CU91-6A }
]
\addplot [forget plot] graphics [xmin=0.5,xmax=191.5,ymin=0.5,ymax=250.5] {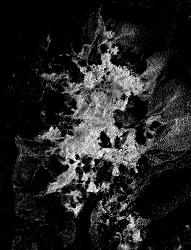};
\end{axis}
\end{tikzpicture}
\begin{tikzpicture}[show background rectangle, tight background, font=\tiny]
\begin{axis}[%
width=\figurewidth,
height=\figureheight,
axis on top,
scale only axis,
xmin=0.5,
xmax=191.5,
y dir=reverse,
ymin=0.5,
ymax=250.5,
hide axis,
name=plot5,
title={$\text{Kaolin}_\text{S}\text{mect H89-FR-5}$} 
]
\addplot [forget plot] graphics [xmin=0.5,xmax=191.5,ymin=0.5,ymax=250.5] {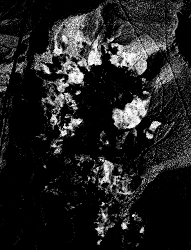};
\end{axis}
\end{tikzpicture}
\begin{tikzpicture}[show background rectangle, tight background, font=\tiny]
\begin{axis}[%
width=\figurewidth,
height=\figureheight,
axis on top,
scale only axis,
xmin=0.5,
xmax=191.5,
ymin=0.5,
ymax=250.5,
hide axis,
title={$\text{Kaolin}_\text{S}\text{mect KLF508}$} 
]
\addplot [forget plot] graphics [xmin=0.5,xmax=191.5,ymin=0.5,ymax=250.5] {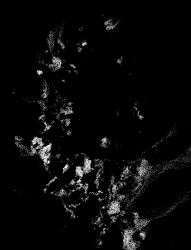};
\end{axis}
\end{tikzpicture}
\\
\begin{tikzpicture}[show background rectangle, tight background, font=\tiny]
\begin{axis}[%
width=\figurewidth,
height=\figureheight,
axis on top,
scale only axis,
xmin=0.5,
xmax=191.5,
y dir=reverse,
ymin=0.5,
ymax=250.5,
hide axis,
name=plot7,
title={Dumortierite HS190.3B }
]
\addplot [forget plot] graphics [xmin=0.5,xmax=191.5,ymin=0.5,ymax=250.5] {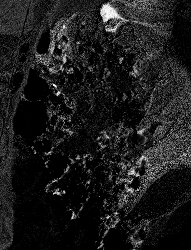};
\end{axis}
\end{tikzpicture}
\begin{tikzpicture}[show background rectangle, tight background, font=\tiny]
\begin{axis}[%
width=\figurewidth,
height=\figureheight,
axis on top,
scale only axis,
xmin=0.5,
xmax=191.5,
y dir=reverse,
ymin=0.5,
ymax=250.5,
hide axis,
name=plot8,
title={Montmorillonite+Illite CM37 }
]
\addplot [forget plot] graphics [xmin=0.5,xmax=191.5,ymin=0.5,ymax=250.5] {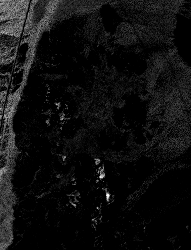};
\end{axis}
\end{tikzpicture}
\begin{tikzpicture}[show background rectangle, tight background, font=\tiny]
\begin{axis}[%
width=\figurewidth,
height=\figureheight,
axis on top,
scale only axis,
xmin=0.5,
xmax=191.5,
y dir=reverse,
ymin=0.5,
ymax=250.5,
hide axis,
name=plot9,
title={Muscovite IL107 }
]
\addplot [forget plot] graphics [xmin=0.5,xmax=191.5,ymin=0.5,ymax=250.5] {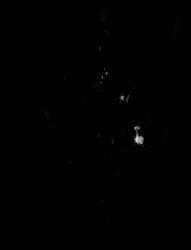};
\end{axis}
\end{tikzpicture}
\\
\begin{tikzpicture}[show background rectangle, tight background, font=\tiny]
\begin{axis}[%
width=\figurewidth,
height=\figureheight,
axis on top,
scale only axis,
xmin=0.5,
xmax=191.5,
y dir=reverse,
ymin=0.5,
ymax=250.5,
hide axis,
name=plot10,
title={Nontronite NG-1.a }
]
\addplot [forget plot] graphics [xmin=0.5,xmax=191.5,ymin=0.5,ymax=250.5] {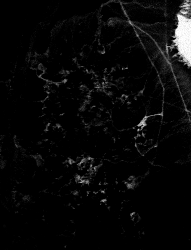};
\end{axis}
\end{tikzpicture}
\begin{tikzpicture}[show background rectangle, tight background, font=\tiny]
\begin{axis}[%
width=\figurewidth,
height=\figureheight,
axis on top,
scale only axis,
xmin=0.5,
xmax=191.5,
y dir=reverse,
ymin=0.5,
ymax=250.5,
hide axis,
name=plot11,
title={Andradite WS474 }
]
\addplot [forget plot] graphics [xmin=0.5,xmax=191.5,ymin=0.5,ymax=250.5] {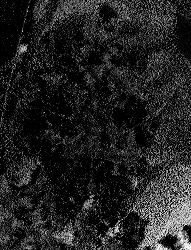};
\end{axis}
\end{tikzpicture}
\begin{tikzpicture}[show background rectangle, tight background, font=\tiny]
\begin{axis}[%
width=\figurewidth,
height=\figureheight,
axis on top,
scale only axis,
xmin=0.5,
xmax=191.5,
y dir=reverse,
ymin=0.5,
ymax=250.5,
hide axis,
title={Sphene HS189.3B }
]
\addplot [forget plot] graphics [xmin=0.5,xmax=191.5,ymin=0.5,ymax=250.5] {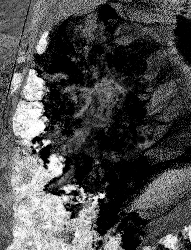};
\end{axis}
\end{tikzpicture}%
   	\caption{Estimated abundance fraction maps for Cuprite dataset}
   	\label{fig_cupriteAbundances}
   	\vspace{-1.5em}
   \end{figure}

The proposed method has been applied to this dataset and results are compared with  VCA \cite{TGRS05_VCA} , SISAL \cite{SISAL} and $L_{1/2}$-NMF \cite{MLNMF:TGRS11_L05NMF} in terms of rmsSAD and presented in Table \ref{table:cupriteresultstable}. Also in Fig. \ref{fig_cupriteSignatures} and \ref{fig_cupriteAbundances} spectral signatures and abundance fraction maps that are extracted using the proposed method are illustrated. Results confirms that this method can effectively unmix hyperspectral data.


\begin{table}[!t]
	\renewcommand{\arraystretch}{1.3}
	\caption{Comparison between methods in terms of rmsSAD for Cuprite Nevada dataset}
	\label{table:cupriteresultstable}
	\centering
	\begin{tabular}{|c|cccc|}
		\cline{1-5}
		Method & VCA & SISAL & $L_{1/2}$-NMF & MLNMF\\
		\hline
		rmsSAD & 0.1047 & 0.1237 & 0.1138 &	\textbf{0.0981}\\
		\hline
	\end{tabular}
\end{table}

\section{Conclusion}
\label{sec:conclusion}
Hyperspectral data contains mixed pixel due to low spatial resolution of the sensors. Spectral unmixing tries to decompose a mixed pixel into the spectral signature and abundance fractions. NMF methods has been widely used to solve spectral unmixing of hyperspectral data. In this paper to improve the performance of NMF method multilayer structure is used. Sparsity constraints on spectral signatures and abundance fractions have been added to the cost function. Synthetic and real datasets are used to evaluate the performance of the proposed mathod in comparison with other methods. Results in terms of SAD and AAD shows that the proposed method can effectively unmix the hyperspectral data.

%


\bibliographystyle{IEEEbib}
\bibliography{strings,refs}

\end{document}